\documentclass[letterpaper, 10 pt, conference]{ieeeconf/ieeeconf}

\IEEEoverridecommandlockouts

\usepackage[utf8]{inputenc}
\usepackage[T1]{fontenc}

\usepackage{balance}  
\usepackage{graphicx}  
\usepackage{booktabs}  
\usepackage{multirow}  
\usepackage{pifont}    
\usepackage{amsmath}
\usepackage{amssymb}
\usepackage{amsfonts}
\usepackage{float}
\usepackage{caption}
\usepackage{xcolor}
\usepackage[capitalise,noabbrev]{cleveref}
\newcommand{\ourshort}{ADAPT}
\newcommand{\XSolidBrush}{\ding{55}}
\newcommand{\Checkmark}{\ding{51}}
\newcommand{\ci}[1]{\tiny{\textcolor{gray}{~($\pm #1$)}}}
\title{\LARGE \bf
ADAPT: Adaptive Dual-projection Architecture for Perceptive Traversal 
}

\author{Shuo Shao$^{1*}$, Tianchen Huang$^{2*}$, Wei Gao$^{1}$\textsuperscript{\textdagger} and Shiwu Zhang$^{1}$\textsuperscript{\textdagger}
\thanks{* denotes equal contribution}%
\thanks{\textdagger \ denotes the corresponding author}%
\thanks{Shuo Shao is with the Department of Automation, University of Science and Technology of China, Anhui 230026, China. {\tt\footnotesize sure318@mail.ustc.edu.cn}}
\thanks{Tianchen Huang, Wei Gao and Shiwu Zhang are with the Institute of Humanoid Robots, Department of Precision Machinery and Precision Instrumentation, University of Science and Technology of China, Hefei, Anhui 230026, China. {\tt\footnotesize weigao@ustc.edu.cn}}%
}

\begin{document}

\maketitle
\thispagestyle{empty}
\pagestyle{empty}

\begin{abstract}
Agile humanoid locomotion in complex 3D environments requires balancing perceptual fidelity with computational efficiency, yet existing methods typically rely on rigid sensing configurations. We propose \textbf{ADAPT} (\textbf{A}daptive \textbf{d}ual-projection \textbf{a}rchitecture for \textbf{p}erceptive \textbf{t}raversal), which represents the environment using a horizontal elevation map for terrain geometry and a vertical distance map for traversable-space constraints. ADAPT further treats its spatial sensing range as a learnable action, enabling the policy to expand its perceptual horizon during fast motion and contract it in cluttered scenes for finer local resolution. Compared with voxel-based baselines, ADAPT drastically reduces observation dimensionality and computational overhead while substantially accelerating training. Experimentally, it achieves successful zero-shot transfer to a Unitree G1 Humanoid and significantly outperforms fixed-range baselines, yielding highly robust traversal across diverse 3D environtmental challenges.
\end{abstract}
    
\section{INTRODUCTION}
Deploying humanoid robots in unstructured environments requires high-fidelity spatial understanding. However, onboard computing imposes strict latency bounds that challenge complex perception systems. The central bottleneck in current legged locomotion research arises from a rigid dichotomy between geometric completeness and computational feasibility. Conventional 2.5D elevation maps~\cite{long2025learning,wang2025beamdojo,ren2025vb,he2025attention,videomimic,Multi-Layer_ElevationMap_2025,Roth_2025_LearnedPerceptiveForwardDynamicsModel,Zhang_2026_AME2,ElevationMap_GPU_2022,Fankhauser2018ProbabilisticTerrainMapping,Fankhauser2014RobotCentricElevationMapping} offer efficiency but remain structurally blind to overhead impediments. Conversely, depth-based representations~\cite{zhuang2024humanoidRobot,zhuang2024humanoid,zhu2026hiking,Yu_2025_START,Castillo_2025_LearningTerrainAwareBipedalLocomotion,Solano_2025_StandWalkNavigate,Sun_2025_DPL,Wang_2025_LocoMamba,Li_2024_MOVE} struggle to fully represent the 3D spatial environment and are highly susceptible to variations in ambient lighting. Volumetric approaches~\cite{Ben_2025_Gallant,Niijima_2025_Real-timeMulti-PlaneSegmentation,Wang_2025_Omni-Perception} capture richer context. However, such methods often suffer from significant processing latency and sparse structural definition.

A more fundamental limitation of existing perception paradigms lies in their reliance on static sensing configurations. These systems typically operate independently of the robot's holistic state. They neglect critical factors such as internal kinematics, velocity and immediate environmental complexity. By decoupling perception from the broader context of action, such approaches impose a rigid and non-adaptive information pipeline. This inflexibility leads to inefficient resource allocation during simple traversal. More critically, it risks catastrophic under-sampling during high-speed maneuvers.

To resolve this conflict, we propose \textbf{ADAPT} (\textbf{A}daptive \textbf{d}ual-projection \textbf{a}rchitecture for \textbf{p}erceptive \textbf{t}raversal). The ADAPT framework restructures 3D environmental data into two orthogonal and information-dense representations. The first is a horizontal elevation map characterizing ground geometry, essential for stable foothold selection. The second is a vertical distance map delineating safe traversable space, crucial for avoiding overhangs and obstacles. Unlike conventional methods that rely on a fixed perceptual horizon, ADAPT elevates the sensing radius to a dynamic control variable learned directly by the reinforcement learning policy. This integration enables the agent to autonomously govern its information intake. The policy proactively expands the perceptual horizon during high-speed transit. Conversely, it narrows the focus to enhance local resolution during intricate maneuvering.

Empirical evaluation demonstrates that this active perception strategy enhances system robustness. ADAPT achieves a 94.7\% success rate across a spectrum of 3D negotiation tasks and surpasses fixed-range baselines by 30\%. The dual-projection architecture reduces observation dimensionality by 98.6\% and lowers training time per iteration by 3.06$\times$ compared to voxel-based alternatives. The learned adaptive sensing policy further reduces computational overhead by 56\% through dynamically matching resource consumption to the complexity of the motion.

The main contributions of this work are threefold:
\begin{itemize}
    \item A novel lightweight dual-projection perception framework that decomposes 3D environments into orthogonal maps.
    \item Introducing an RL-driven adaptive sensing mechanism that treats the sensing radius as a control parameter to autonomously balance speed and safety.
    \item Robust zero-shot sim-to-real transfer on a Unitree G1 humanoid with successful navigation through diverse structural challenges.
\end{itemize}

\section{RELATED WORK}
\subsection{Perception for Legged Locomotion}
\label{sec:RELATED WORKS}
Perception systems for legged locomotion historically face a trade-off between geometric fidelity and computational efficiency, as summarized in Table~\ref{table:comparison_method}. Elevation Map-based methods~\cite{long2025learning,wang2025beamdojo,ren2025vb,Fankhauser2018ProbabilisticTerrainMapping,Fankhauser2014RobotCentricElevationMapping} utilize 2.5D elevation grids to enable rapid foothold planning. However, the 2.5D structure fails to represent overhanging obstacles and creates blind spots in complex environments. Alternatively, depth-based approaches~\cite{zhuang2024humanoid,Sun_2025_DPL} capture scene geometry directly but lack explicit 3D spatial semantics and remain vulnerable to illumination changes. To address dimensionality, volumetric representations~\cite{Ben_2025_Gallant,Niijima_2025_Real-timeMulti-PlaneSegmentation,Xue_2026_Collision-FreeHumanoidTraversal,Wang_2025_Omni-Perception} provide full 3D awareness. Yet, the cubic complexity of voxel grids forces a compromise between usable resolution and system latency. Moreover, high computational overhead increases processing latency during high-speed maneuvers. This delay can worsen state-estimation drift and reduce control stability. These limitations motivate the development of lightweight and low-latency perceptual representations in this work.

A polarization persists in modern environment representation. Simplified 2.5D elevation map sacrifices vertical context and ignores overhead obstacles. Conversely, dense volumetric grids provide comprehensive details but suffer from massive data redundancy. Such redundancy hinders reinforcement learning convergence. ADAPT resolves this dilemma by restructuring the 3D environment into two orthogonal projections. We decouple the environment representation into a horizontal map for terrain geometry and a vertical map for traversable space analysis. This dual-map architecture captures comprehensive 3D semantics, including obstacle proximity. The approach maintains a lightweight observation space and effectively combines volumetric fidelity with the efficiency of 2D maps.

\begin{table}[t]
    \centering
    \vspace{-5pt}
    \caption{ADAPT comparison with existing works}
    \resizebox{\columnwidth}{!}{
    \begin{tabular}{lccccc}
    \toprule 
    \multirow{2}{*}{Method} & \multirow{2}{*}{\shortstack{Perceptual\\Representation}} & \multirow{2}{*}{\shortstack{Native\\Sampling}} & \multirow{2}{*}{\shortstack{Sensory\\Dimension}} & \multirow{2}{*}{\shortstack{Overhead}} & \multirow{2}{*}{\shortstack{Adaptive\\Sensing}} \\ \\
    \midrule
    PIM\cite{long2025learning} & Elevation Map & 0.10 m & 96 & \XSolidBrush & \XSolidBrush \\
    Beamdojo\cite{wang2025beamdojo} & Elevation Map & 0.10 m & 225 & \XSolidBrush & \XSolidBrush \\
    Vb-com\cite{ren2025vb} & Elevation Map & 0.10 m & 225 & \XSolidBrush & \XSolidBrush \\ 
    Parkour\cite{zhuang2024humanoid} & Depth Image & 48$\times$64 px & 3072 & \Checkmark & \XSolidBrush \\ 
    Omni\cite{Wang_2025_Omni-Perception} & Point Cloud & 0.05$\pi$ rad & 251 & \Checkmark & \XSolidBrush \\ 
    Gallant\cite{Ben_2025_Gallant} & Voxel Grid & 0.05 m & 40960 & \Checkmark & \XSolidBrush \\
    HumanoidPF\cite{Xue_2026_Collision-FreeHumanoidTraversal} & Joint Position & 39 dims & 39 & \Checkmark & \XSolidBrush \\
    \cmidrule(r){1-6}
    \ourshort{} (ours) & Dual Projection & \textbf{0.025$\pi$ rad} & \textbf{578} & \Checkmark & \Checkmark \\
    \bottomrule
    \end{tabular}}
    \label{table:comparison_method}
    \vspace{-10pt}
\end{table}
\subsection{Adaptive Perception and Behavior Modulation}
The tension between perceptual fidelity and computational cost has driven research into adaptive sensing strategies. Some works have explored spatial selective attention. For instance, Wang et al.~\cite{Wang_2025_AdaptiveVision} use reinforcement learning to direct foveation toward task-relevant sub-regions. Similarly, Zhang et al.~\cite{Zhang_2026_FocusNav} employ a stability-aware gating mechanism to filter distal noise during unstable locomotion. Parallel research investigates hierarchical data representations. Patel et al.~\cite{Patel_2024_RoadRunnerMM} and Liu et al.~\cite{Liu_2026_ActiveVLA} utilize predefined multi-resolution tiers or active zoom mechanisms to adjust data density dynamically.

Most existing adaptive methods depend on discrete and manually tuned heuristics. Such reliance creates a loose coupling between perception and actuation. In contrast, ADAPT establishes a bidirectional feedback loop where the sensing radius functions as both a control action and a state observation. Through end-to-end reinforcement learning, the robot acquires an emergent capability to couple the perceptual horizon with instantaneous velocity and environmental complexity. This mechanism allows the system to optimize the signal-to-noise ratio autonomously. Consequently, the policy ensures safe negotiation of hazardous terrain and ignores irrelevant distal features.

\section{METHOD}
\subsection{Problem Formulation}
Navigating unstructured environments requires fusing proactive terrain perception with whole-body motion adaptation. We formalize this traversal task as a goal-conditioned Partially Observable Markov Decision Process (POMDP) defined by the tuple $(\mathcal{S}, \mathcal{O}, \mathcal{A}, \mathcal{T}, \mathcal{R}, \gamma)$.
At each time step $t$, the agent receives a partial observation $o_t \in \mathcal{O}$ derived from the latent state $s_t \in \mathcal{S}$. This observation encapsulates the robot's kinematic configuration and local environmental context. The policy $\pi_\theta: \mathcal{O} \rightarrow \mathcal{A}$ generates a composite action $a_t = (q_t, r_t) \in \mathcal{A}$ that comprises joint position targets $q_t$ and an adaptive perception radius $r_t$. This dual-action structure simultaneously regulates physical actuation and sensory intake. The environment evolves according to stochastic dynamics $\mathcal{T}: \mathcal{S} \times \mathcal{A} \rightarrow \Delta(\mathcal{S})$ governed by the physics simulator. The agent maximizes a scalar reward $R_t = \mathcal{R}(s_t, a_t)$ encoding objectives spanning velocity tracking, energy efficiency and safety constraints.

The learning objective optimizes the policy $\pi^*$ to maximize the expected cumulative discounted return defined as
\begin{equation}
    \pi^* = \arg\max_{\pi} \mathbb{E}_{\pi} \left[ \sum_{t=0}^{T} \gamma^t \mathcal{R}(s_t, a_t) \right]
\end{equation}
where $\gamma \in (0, 1)$ represents the discount factor and $T$ denotes the horizon.

\subsection{Training Pipeline}
\subsubsection{Observation Space}
\label{sssec:obs}

\begin{figure*}[t]
    \centering
    \includegraphics[width=0.95\textwidth]{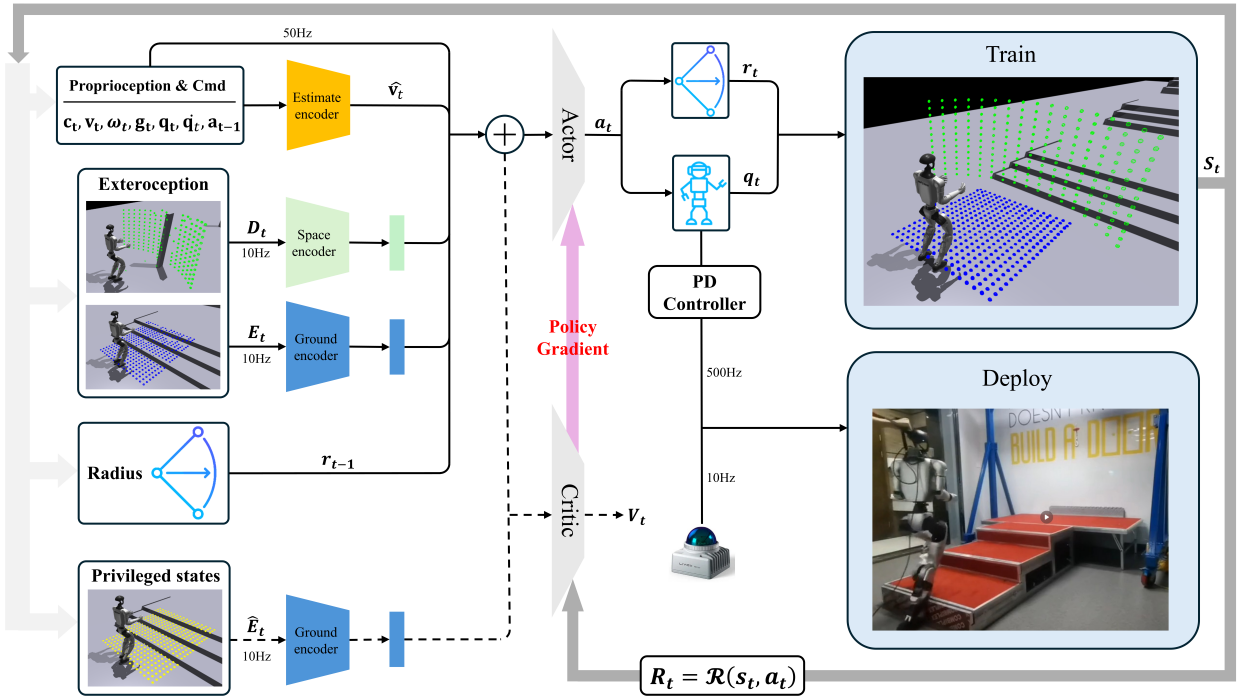}
    \caption{The proposed training architecture separates exteroceptive and proprioceptive inputs. Two MLP encoders process the perception maps independently. Their features are then fused with the proprioceptive state and passed to a GRU-based recurrent actor-critic. The actor predicts both the joint targets and the perception radius for the next step. The critic uses privileged, noise-free simulator information to produce a more accurate value estimate and stabilize training.}
    \label{fig:network_architecture}
\end{figure*}

\begin{figure}[h]
    \centering
    \includegraphics[width=0.9\columnwidth]{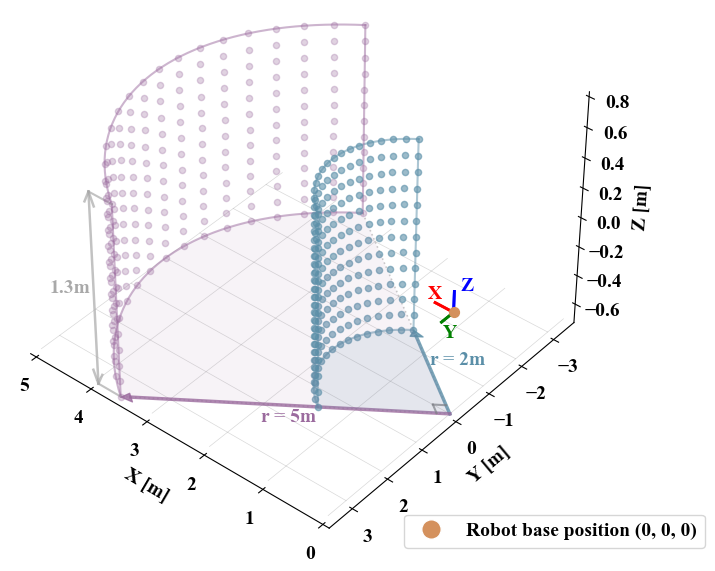}
    \caption{Illustration of the adaptive perception radius mechanism. The vertical distance map is constructed on a cylindrical sector with a fixed number of cells ($17 \times 13$). A smaller radius (blue) provides higher spatial resolution for precise near-field sensing, while a larger radius (purple) extends the perception range at the cost of reduced resolution. 
    The policy learns to dynamically adjust this radius based on the robot's locomotion state.}
    \label{fig:adaptive_radius}
\end{figure}

The observation vector $o_t$ integrates proprioceptive feedback, task commands and exteroceptive data. It is defined as
\begin{equation}
\mathbf{o}_t = \mathrm{concat}\Big(
\mathbf{c}_t, 
\mathbf{v}_t, \boldsymbol{\omega}_t, \mathbf{g}_t, \mathbf{q}_t, \dot{\mathbf{q}}_t, \mathbf{a}_{t-1},
\mathbf{E}_t, \mathbf{D}_t, r_{t-1}
\Big).
\label{eq:observation_vector}
\end{equation}
The specific components in~\eqref{eq:observation_vector} are:
\begin{itemize}
    \item \textbf{Task Command:} The target velocity command $\mathbf{c}_t \in \mathbb{R}^3$.
    \item \textbf{Proprioceptive States:} These states include base linear velocity $\mathbf{v}_t$, base angular velocity $\boldsymbol{\omega}_t$, the gravitational vector $\mathbf{g}_t$, joint positions $\mathbf{q}_t$, joint velocities $\dot{\mathbf{q}}_t$ and the preceding action $\mathbf{a}_{t-1}$.
    \item \textbf{Exteroceptive States:} The actor receives synthetic point cloud data during learning, acquired via the LiDAR simulation framework from Wang et al.~\cite{Wang_2025_Omni-Perception}. 
    We rasterize the raw point cloud into grid structures to facilitate efficient processing. Consequently, this input is processed into two orthogonal robot-centric projections as shown in Fig.~\ref{fig:network_architecture}:
    \begin{itemize}
        \item \textit{Horizontal Elevation Map ($\mathbf{E}_t \in \mathbb{R}^{21 \times 17}$):} Captures terrain topology via discrete height sampling.
        \item \textit{Vertical Distance Map ($\mathbf{D}_t \in \mathbb{R}^{13 \times 17}$):} Encodes spatial traversability. We generate this map by projecting points into a polar sector spanning from -45 to +45 degrees around the vertical axis, where each cell records the distance to the nearest obstacle.
    \end{itemize}
    \item \textbf{Adaptive Perception Radius:} The radius $r_{t-1}$ acts as a dynamic gain modulating the field of view for both projection maps. We constrain the sensing radius to $r_t \in [1, 5]$\,m throughout training and deployment. The vertical distance map uses a cylindrical sector grid with a fixed number of bins $(N_r, N_\theta) = (13, 17)$ as illustrated in Fig.~\ref{fig:adaptive_radius}. For a selected radius $r_{t-1}$, the radial sampling interval is defined as
    \begin{equation}
        \delta_r = \frac{r_{t-1}}{N_r},
    \end{equation}
    which makes the physical cell size grow linearly with the sensing radius. Therefore, a smaller radius concentrates the same grid on the immediate vicinity and yields finer near-field resolution. Additionally, limiting the range reduces computational overhead and minimizes distal sensor noise. Conversely, a larger radius extends the perceptual horizon but necessarily coarsens local sampling resolution.
\end{itemize}
    
The critic operates on a privileged observation space to facilitate accurate estimation of the value function $V_t$. In addition to the actor's input vector $o_t$, the critic receives the noise-free, ground-truth elevation map $\hat{\mathbf{E}}_t$ directly acquired from the simulation engine.
We scale, flatten and concatenate all components into a 648-dimensional vector serving as the input to the recurrent policy.

\subsubsection{Action Space}
The policy produces a composite action $\mu_t \in \mathbb{R}^{13}$ for each step at $50$ Hz to simultaneously handle motor control and perceptual modulation. 
The composite action $a_t$ is thus formally defined as:
\begin{equation}
    a_t = \{q_t, r_t\}
\end{equation}

\begin{itemize}
    \item \textbf{Joint Targets:} The initial 12 elements of $\mu_t$ specify target joint positions. We scale and superimpose these targets onto a nominal configuration $q_{\text{default}}$ to yield the actionable targets $q_t \in \mathbb{R}^{12}$ defined as
    \begin{equation}
        q_t = s_a \cdot \mu_{t, 1:12} + q_{\text{default}}
    \end{equation}
    where $s_a$ is the action scale. 
    These targets are executed by a low-level Proportional-Derivative (PD) controller.
    \item \textbf{Adaptive Perception Radius:} The final element of $\mu_t$ dictates the perception radius $r_t \in [1, 5]$\,m for the subsequent timestep. This structure establishes a feedback loop wherein the policy proactively shapes its future informational state based on current locomotion dynamics.
\end{itemize}

\subsubsection{Reward Functions}
The reward function $\mathcal{R}$ integrates multiple objectives to encourage robust and efficient locomotion adhering to safety constraints. The total reward $r_t = \sum_i w_i r_i(t)$ is a weighted sum of the following components:

\begin{itemize}
    \item \textbf{Task Fulfillment:} We encourage precise command tracking by rewarding close alignment with target linear and angular velocities:
    \begin{equation}
    \begin{split}
        r_{\text{track}} ={}& w_{\text{lin}} \exp(-||\mathbf{v}_t - \mathbf{c}_{t, \text{lin}}||^2) \\
                        & + w_{\text{ang}} \exp(-||\omega_{z,t} - c_{t, \text{yaw}}||^2).
    \end{split}
    \end{equation}
    
    \item \textbf{Locomotion Stability:} We maintain postural integrity by penalizing deviations in roll and pitch ($r_{\text{orient}}$), excessive base angular velocities in the horizontal plane ($r_{\text{ang\_vel\_xy}}$) and hip joint divergence from neutral configurations ($r_{\text{hip}}$).
    
    \item \textbf{Behavior Regularization:} We promote energy efficiency and motion smoothness by penalizing high joint velocities ($r_{\text{dof\_vel}}$), accelerations ($r_{\text{dof\_acc}}$) and high-frequency action oscillations combined with torque squared minimization ($r_{\text{energy}}$).
    
    \item \textbf{Safety and Survival:} We discourage critical failures via penalties for body collisions ($r_{\text{collision}}$), mesh interpenetration ($r_{\text{penetrate}}$) and joint limit violations ($r_{\text{dof\_limits}}$). A continuous survival bonus ($r_{\text{alive}}$) incentivizes prolonged operation while foot air-time rewards ($r_{\text{air\_time}}$) encourage the dynamic gait characteristics necessary for clearing obstacles.

    \item \textbf{Adaptive Perception Rewards:} We shape active sensing using three specialized terms, with their weights empirically tuned to balance their competing objectives:
    \begin{itemize}
        \item \textit{Adaptive Tracking ($r_{\text{radius\_adaptive}}$)} aligns the predicted radius with a targeted optimal value via a Gaussian function. This desired radius is dynamically computed based on the robot's forward speed and proximity to the nearest obstacle (extracted from $\mathbf{D}_t$). High speeds and open spaces encourage a larger radius (up to $5$\,m), whereas slow speeds or nearby clutter dictate a smaller radius (down to $1$\,m) to maximize near-field resolution.
        \item \textit{Smoothness Penalty ($r_{\text{radius\_smoothness}}$)} penalizes abrupt frame-to-frame variations to prevent oscillating sensing configurations.
        \item \textit{Boundary Regularization ($r_{\text{radius\_regularization}}$)} discourages the commanded radius from continuously saturating at its physical limits.
    \end{itemize}
\end{itemize}

\subsubsection{Terrain Design}

\begin{figure}[t]
    \centering
    \includegraphics[width=\columnwidth]{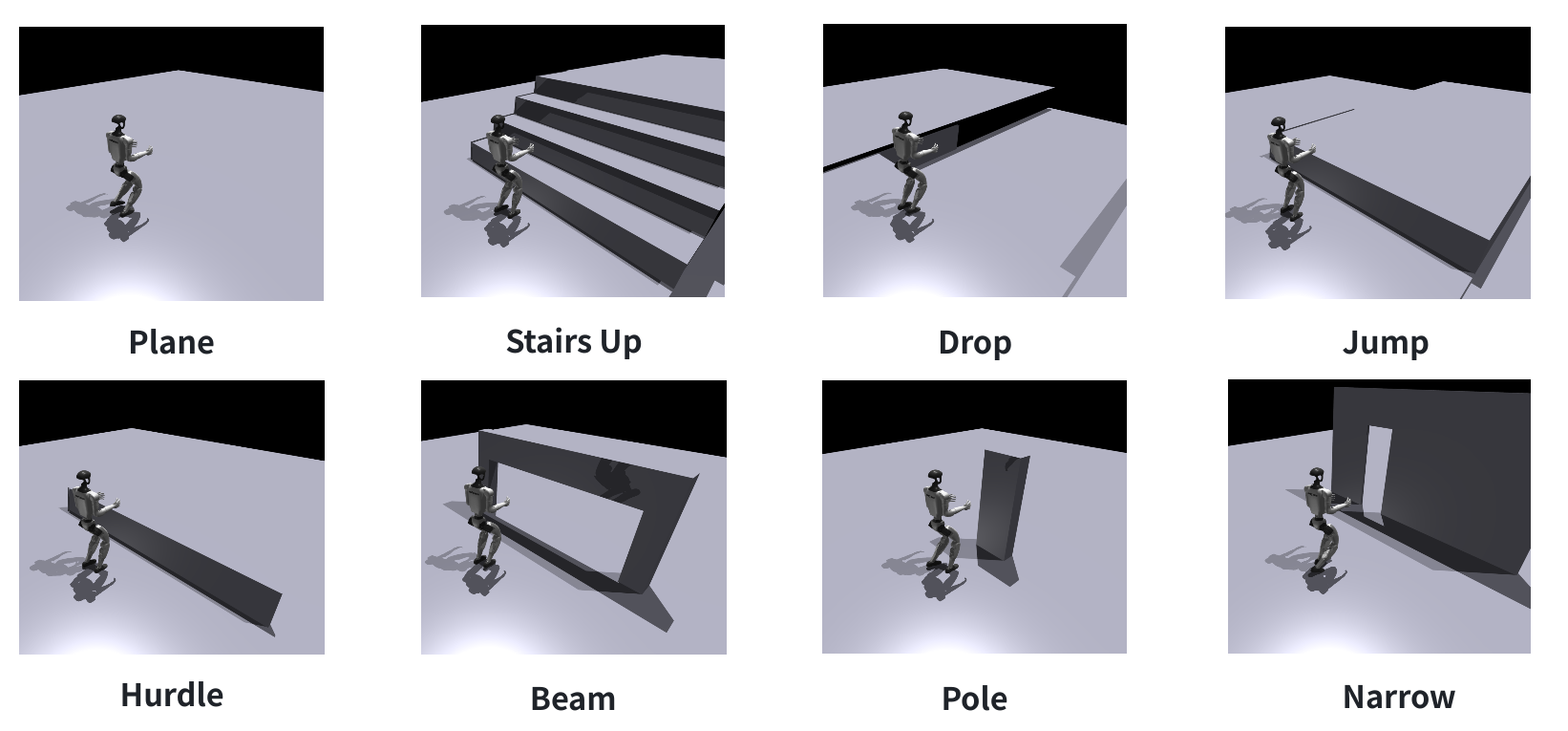}
    \caption{Examples of procedurally generated obstacles in our \textit{BarrierTrack} training environment. The curriculum includes a diverse set of challenges such as stairs, gaps, hurdles, and constrained passages like beams and narrow gates, promoting the development of a versatile locomotion policy.}
    \label{fig:terrain}
\end{figure}

We employ our \textit{BarrierTrack} environment to cultivate a versatile policy. This system constructs diverse parkour courses by sequentially assembling discrete obstacle modules. This sequential assembly subjects the agent to a spectrum of geometric challenges.
The training process utilizes a curriculum strategy scaling obstacle difficulty in correlation with policy performance. Each episode features a unique track generated by permuting fundamental obstacle primitives quantified in Table~\ref{tab:terrain_params}.
The obstacle dimensions are sampled from the continuous ranges to ensure intra-class variability. Furthermore, Perlin noise is superimposed onto ground surfaces to simulate terrain irregularity and prevent overfitting to idealized planar geometry. This synthesis of modular design, curriculum learning and stochastic variation creates a generalizable locomotion policy.

\begin{table}[h]
\centering
\caption{Parameters for generating curriculum training terrains.}
\label{tab:terrain_params}
\setlength{\tabcolsep}{4pt}
\begin{tabular}{@{}llcc@{}}
\toprule
\textbf{Terrain} & \textbf{Param. (m)} & $\mathbf{p}_\tau^{\min}$ & $\mathbf{p}_\tau^{\max}$ \\
\midrule
\multirow{2}{*}{\textbf{Stairs Up}} & Step Height $\uparrow$ & 0.10 & 0.30 \\
& Step Length $\downarrow$ & 0.30 & 0.50 \\
\addlinespace
\multirow{2}{*}{\textbf{Drop}} & Drop Height $\uparrow$ & 0.10 & 0.60 \\
& Platform Length $\downarrow$ & 0.30 & 0.50 \\
\addlinespace
\multirow{2}{*}{\textbf{Jump}} & Wall Height $\uparrow$ & 0.05 & 0.40 \\
& Wall Depth $\uparrow$ & 0.10 & 0.30 \\
\addlinespace
\multirow{2}{*}{\textbf{Hurdle}} & Hurdle Height $\uparrow$ & 0.05 & 0.40 \\
& Hurdle Depth $\uparrow$ & 0.05 & 0.10 \\
\addlinespace
\multirow{2}{*}{\textbf{Beam}} & Clearance Height $\downarrow$ & 1.33 & 1.10 \\
& Beam Depth $\uparrow$ & 0.20 & 0.20 \\
\addlinespace
\multirow{2}{*}{\textbf{Pole}} & Pole Width $\downarrow$ & 0.50 & 0.10 \\
& Position Offset $\uparrow$ & 0.00 & 0.00 \\
\addlinespace
\multirow{2}{*}{\textbf{Narrow Gate}} & Gate Width $\downarrow$ & 0.80 & 0.45 \\
& Position Offset $\uparrow$ & 0.00 & 0.60 \\
\bottomrule
\end{tabular}
\end{table}


\subsection{Training Architecture}
\label{sec:training_architecture}

We optimize the policy using the Proximal Policy Optimization (PPO) algorithm. The training framework employs an asymmetric recurrent actor-critic architecture with a Gated Recurrent Unit (GRU) to capture short-horizon temporal dependencies during locomotion. This framework allows the critic to leverage privileged simulation states to accelerate convergence. 
The critic combines the recurrent state with noise-free ground truth including true elevation to enable precise value estimation. 
In contrast, the actor operates exclusively on deployable observations.
This asymmetry facilitates robust policy learning while ensuring the actor remains deployable under real-world sensor limitations.

The policy network processes the observation vector $o_t$ (Sec.~\ref{sssec:obs}) through the segmented architecture shown in Fig.~\ref{fig:network_architecture}. Two separate MLP encoders independently encode the exteroceptive inputs $\mathbf{E}_t$ and $\mathbf{D}_t$ into 32-dimensional latent variables. These latent representations are concatenated to form a 64-dimensional environmental embedding and fused with the proprioceptive state before being passed to the GRU-based actor and critic heads.

A distinguishing feature of the proposed architecture is the closed-loop modulation of perceptual parameters. Conditioned on the recurrent hidden state, the actor outputs a 13-dimensional vector encompassing joint targets and the perception radius $r_t$. This design tightly couples the sensing range with the robot's kinematic states and environmental proximity to effectively treat perception as an actuated degree of freedom.



\section{Experiment}
\label{sec:experiment}

\subsection{Experimental Configuration}
\label{sec:method_terrain}

\subsubsection{Simulation Environment} All policies are trained using NVIDIA Isaac Gym~\cite{Isaac_Gym} on an NVIDIA RTX 4090 GPU with the 12-DoF Unitree G1 humanoid robot. LiDAR point clouds are synthesized using the simulator from~\cite{Wang_2025_Omni-Perception}. The simulator accurately models sensor noise along with occlusion from static terrain and dynamic robot links.

\subsubsection{Real-world Environment} The trained policy is deployed on the physical Unitree G1. A head-mounted Livox Mid-360 LiDAR provides raw point cloud data. State estimation and point cloud registration are handled by Fast-LIO2~\cite{Xu_2022_FAST-LIO2}. The horizontal elevation map $\mathbf{E}_t$ is extracted from the registered point cloud using the GPU-accelerated elevation mapping pipeline from~\cite{ElevationMap_GPU_2022} at 10\,Hz. The vertical distance map $\mathbf{D}_t$ is generated at 10\,Hz. Registered points are projected into a polar sector spanning $-45^\circ$ to $+45^\circ$. Each map cell records the distance to the nearest obstacle. The map generation follows the exact same procedure used in simulation. Joint commands are executed at 50\,Hz on an onboard NVIDIA Jetson Orin NX.

\subsection{Simulation Experiments}
\subsubsection{Evaluation Setup}
All methods are evaluated on the hardest difficulty tier of \textit{BarrierTrack}. The benchmark includes seven terrain archetypes, grouped into two categories: precise foot placement with Stairs, Drop, Jump and Hurdle, and traversable-space selection with Beam, Pole and Narrow Gate. For each terrain type, every episode contains five obstacles with progressively increasing difficulty; obstacle parameters are sampled within [$\mathbf{p}_\tau^{\min}$, $\mathbf{p}_\tau^{\max}$] according to the curriculum schedule (Table~\ref{tab:terrain_params}). This design exposes distinct locomotion sub-skills while enforcing within-episode difficulty escalation. A random sequence of terrain modules is assembled for each episode, forcing the agent to solve compound geometric challenges rather than isolated obstacles.

\subsubsection{Metrics}
Performance is evaluated by the mean and standard deviations over two complementary metrics:
\begin{itemize}
    \item \textbf{Success Rate ($R_{\mathrm{succ}}$)} represents the fraction of discrete obstacles successfully navigated within a single trial. It measures the policy's obstacle-negotiation reliability.
    \item \textbf{Normalized Traversal Progress ($R_{\mathrm{trav}}$)} represents the forward distance traveled along the course normalized by the total track length. It evaluates the overall capability to string together continuous locomotion skills.
\end{itemize}

\subsubsection{Performance Comparison}
ADAPT (Ours) is benchmarked against four baselines to evaluate its navigation capability and computational efficiency.
Each method is evaluated across all seven terrain types at three commanded locomotion speeds: $v \in \{0.5, 1.0, 1.5\}$\,m/s.
For each speed–terrain combination, 10 trials are conducted and the results are averaged to produce the final reported $R_{\mathrm{succ}}$.
The four baselines include:
\begin{itemize}
    \item \textbf{PIM~\cite{long2025learning}:} A single-stage training method that uses a LiDAR point cloud elevation map as exteroceptive input. Contrastive learning aligns the generated elevation map with ground truth.
    \item \textbf{Parkour~\cite{zhuang2024humanoid}:} A two-stage method. First, a teacher policy is trained with ground-truth elevation maps. Then, supervised distillation transfers agile locomotion skills to a student policy that uses only depth images.
    \item \textbf{TSDF:} A camera-based approach that performs 3D reconstruction via nvblox~\cite{nvblox_2023_GPUAcceleratedSDFMapping} and feeds local grid TSDF values as exteroceptive input for policy training.
    \item \textbf{Gallant~\cite{Ben_2025_Gallant}:} A voxel-based approach that provides full 3D awareness but at a higher computational cost.
\end{itemize}

\begin{table*}[t]
\caption{Benchmarked Comparison in Simulation. $R_\mathrm{succ}$: success rate (\%, $\uparrow$). $R_\mathrm{trav}$: normalized traversal progress (\%, $\uparrow$).}
\label{tab:main_results}
\setlength{\tabcolsep}{4pt}
\begin{center}
\begin{tabular}{lcccccccc}
\toprule[1.0pt]
\multirow{2}{*}{Method} &
\multicolumn{2}{c}{Plane} & \multicolumn{2}{c}{Stairs} & \multicolumn{2}{c}{Drop} & \multicolumn{2}{c}{Jump} \\
\cmidrule(lr){2-3}\cmidrule(lr){4-5}\cmidrule(lr){6-7}\cmidrule(lr){8-9}
& $R_\mathrm{succ}$ & $R_\mathrm{trav}$ & $R_\mathrm{succ}$ & $R_\mathrm{trav}$ & $R_\mathrm{succ}$ & $R_\mathrm{trav}$ & $R_\mathrm{succ}$ & $R_\mathrm{trav}$ \\
\midrule[0.8pt]
PIM~\cite{long2025learning}        & 100\ci{0.0} & 100\ci{0.0} & 65.33\ci{37.57} & 70.00\ci{34.61} & 99.33\ci{3.59} & 99.77\ci{1.26} & 90.67\ci{16.92} & 91.67\ci{14.62} \\[0.4ex]
Parkour~\cite{zhuang2024humanoid}  & 100\ci{0.0} & 100\ci{0.0} & 28.00\ci{28.10} & 37.20\ci{24.93} & 91.33\ci{14.31} & 92.40\ci{13.52} & 44.67\ci{20.45} & 51.40\ci{17.49} \\[0.4ex]
TSDF                               & 100\ci{0.0} & 100\ci{0.0} & 40.00\ci{20.00} & 47.43\ci{17.50} & 56.67\ci{24.81} & 63.33\ci{22.75} & 43.33\ci{13.74} & 48.63\ci{13.65} \\[0.4ex]
Gallant~\cite{Ben_2025_Gallant}    & 100\ci{0.0} & 100\ci{0.0} & 70.00\ci{22.95} & 76.30\ci{21.75} & 80.00\ci{31.41} & 83.23\ci{27.62} & 68.00\ci{13.27} & 74.10\ci{11.34} \\[0.4ex]
\textbf{Ours}                      & \textbf{100\ci{0.0}} & \textbf{100\ci{0.0}} & \textbf{94.67\ci{13.60}} & \textbf{95.90\ci{10.63}} & \textbf{100.00\ci{0.00}} & \textbf{100.00\ci{0.00}} & \textbf{98.67\ci{4.99}} & \textbf{98.90\ci{4.12}} \\[0.4ex]
\midrule[0.8pt]
\multirow{2}{*}{Method} &
\multicolumn{2}{c}{Hurdle} & \multicolumn{2}{c}{Beam} & \multicolumn{2}{c}{Pole} & \multicolumn{2}{c}{Narrow Gate} \\
\cmidrule(lr){2-3}\cmidrule(lr){4-5}\cmidrule(lr){6-7}\cmidrule(lr){8-9}
& $R_\mathrm{succ}$ & $R_\mathrm{trav}$ & $R_\mathrm{succ}$ & $R_\mathrm{trav}$ & $R_\mathrm{succ}$ & $R_\mathrm{trav}$ & $R_\mathrm{succ}$ & $R_\mathrm{trav}$ \\
\midrule[0.4pt]
PIM~\cite{long2025learning}        & 82.00\ci{20.88} & 86.37\ci{16.44} & 66.67\ci{9.43} & 75.53\ci{5.28} & 2.67\ci{6.80} & 6.17\ci{7.09} & 27.33\ci{15.04} & 31.17\ci{14.17} \\[0.4ex]
Parkour~\cite{zhuang2024humanoid}  & 47.33\ci{13.15} & 52.67\ci{12.81} & 14.00\ci{11.72} & 20.53\ci{6.93} & 5.33\ci{8.84} & 15.00\ci{7.90} & 48.00\ci{16.81} & 53.03\ci{16.05} \\[0.4ex]
TSDF                               & 49.33\ci{11.23} & 55.47\ci{13.98} & 33.33\ci{11.93} & 40.83\ci{11.37} & 7.33\ci{9.64} & 14.37\ci{9.07} & 22.67\ci{16.11} & 29.80\ci{14.31} \\[0.4ex]
Gallant~\cite{Ben_2025_Gallant}    & 54.67\ci{16.28} & 62.47\ci{14.21} & 32.00\ci{25.61} & 40.40\ci{27.17} & 46.00\ci{44.17} & 49.40\ci{41.82} & 66.00\ci{25.90} & 69.57\ci{23.35} \\[0.4ex]
\textbf{Ours}                      & \textbf{90.00\ci{14.38}} & \textbf{91.80\ci{12.80}} & \textbf{96.67\ci{7.45}} & \textbf{97.57\ci{5.47}} & \textbf{98.00\ci{6.00}} & \textbf{98.57\ci{4.79}} & \textbf{80.00\ci{18.62}} & \textbf{83.03\ci{16.30}} \\[0.4ex]
\bottomrule[1.0pt]
\end{tabular}
\end{center}
\end{table*}

Table~\ref{tab:main_results} details the performance comparison alongside the associated training costs in Table~\ref{tab:efficiency}. The proposed method (\textbf{Ours}) achieves the highest success rates across all terrains while maintaining the lowest training overhead. 
The performance of the baselines is consistent with the sensing characteristics within each method.
PIM relies on a 2.5D elevation map and shows reduced performance on Pole and Narrow Gate terrains, where vertical and lateral clearance cues are important. Parkour uses depth images and performs less reliably on terrains requiring broader spatial coverage, such as Jump, Hurdle and Beam. TSDF provides volumetric awareness, but its performance remains limited on fine-grained terrains although incurring the largest training cost. Gallant improves spatial coverage relative to 2.5D and depth-based baselines, but its higher-dimensional voxel input requires higher computational cost and achieves lower success rates than ADAPT on most terrains. Overall, these results support that ADAPT provides a favorable balance between traversal performance and training efficiency across both foot-placement and traversable-space-selection tasks.

\begin{table}[t]
\caption{Training Efficiency Comparison. Train/iter: wall-clock time per training iteration (s, $\downarrow$). GPU peak: peak GPU memory usage (MB, $\downarrow$).}
\label{tab:efficiency}
\setlength{\tabcolsep}{6pt}
\begin{center}
\begin{tabular}{lcc}
\toprule[1.0pt]
Method & Train/iter (s, $\downarrow$) & GPU Peak (MB, $\downarrow$) \\
\midrule[0.8pt]
Parkour~\cite{zhuang2024humanoid} & 97.8 & 28791 \\[0.4ex]
Gallant~\cite{Ben_2025_Gallant}   & 34.6 & 36054 \\[0.4ex]
TSDF                                     & 246 & 38473 \\[0.4ex]
\textbf{Ours}                     & \textbf{11.3} & \textbf{15870} \\
\bottomrule[1.0pt]
\end{tabular}
\end{center}
\end{table}

\subsubsection{Ablation Study on Adaptive Mechanism}
A comprehensive ablation study examines the adaptive sensing radius and its reward design. 

\paragraph{Experimental Configurations}
To specifically validate the obstacle avoidance capability of the adaptive mechanism, the ablation study focuses on the perception-critical Beam, Pole and Narrow Gate terrains. 
Three distinct settings isolating the effect of radius representation and reward formulation are considered:
\begin{itemize}
    \item \textbf{Fixed Radius:} The sensing radius $r_t$ is excluded from the observation and action spaces. The policy operates with a constant radius. Separate policies are trained with $R \in \{1, 3, 5\}$\,m.
    \item \textbf{w/o Adaptive Reward:} The sensing radius $r_t$ serves as both an observation, enabling closed-loop regulation, and an action, allowing for the adjustment of perception settings. However, this configuration excludes the reward function related to $r_t$.
    \item \textbf{Adaptive Radius (Ours):} This represents the proposed full approach. The radius is dynamically controlled by the adaptive reward.
\end{itemize}

\paragraph{Results and Analysis}
Table~\ref{tab:ablation_radius} summarizes the ablation results. 
A fixed short radius ($R=1$) yields the lowest average success rates, particularly at higher speeds, indicating that the limited look-ahead range is insufficient for these terrains. A fixed long radius ($R=5$) performs better than $R=1$ at high speed but remains clearly below the full adaptive model, suggesting a trade-off between larger look-ahead range and reduced local resolution. A moderate fixed radius ($R=3$) outperforms both extremes, but its performance still drops as commanded speed increases. Removing the adaptive reward (w/o Adap. Reward) leads to a large performance drop at $v=1.5$\,m/s compared with the full model, which indicates that exposing the radius as an action alone is not sufficient to recover the best sensing behavior. Overall, the fully adaptive method achieves the highest average success rate across all tested speeds, indicating the efficacy of closed-loop radius modulation.

\begin{table}[!ht]
\caption{Ablation on adaptive sensing radius. $R_\mathrm{succ}$: success rate (\%, $\uparrow$). Values are averaged over Beam, Pole, and Narrow Gate terrains.}
\label{tab:ablation_radius}
\begin{center}
\resizebox{\columnwidth}{!}{%
\begin{tabular}{llccc}
\toprule[1.0pt]
Type & Config & $v{=}0.5$\,m/s & $v{=}1.0$\,m/s & $v{=}1.5$\,m/s \\
\midrule[0.8pt]
\multirow{3}{*}{Fixed Radius}  & $R = 1$               & 51.33\ci{17.65} & 51.33\ci{23.49} & 24.67\ci{22.32} \\[0.4ex]
                               & $R = 3$               & 81.33\ci{21.25} & 72.67\ci{15.90} & 56.67\ci{29.70} \\[0.4ex]
                               & $R = 5$               & 64.00\ci{18.90} & 64.67\ci{26.17} & 53.33\ci{30.26} \\
\midrule[0.5pt]
w/o Adap.\ Reward              & no $r_{\text{adap}}$  & 92.67\ci{10.93} & 71.33\ci{25.66} & 42.67\ci{29.09} \\
\midrule[0.5pt]
Adaptive Radius                & \textbf{Ours}         & \textbf{94.67\ci{10.24}} & \textbf{96.00\ci{8.00}} & \textbf{84.00\ci{19.60}} \\
\bottomrule[1.0pt]
\end{tabular}%
}
\end{center}
\end{table}

\subsection{Discussion}

The consistent performance in simulation and the real world shows the simplicity and effectiveness of ADAPT. Unlike complex baselines with heavy 3D reconstruction or multi-stage training, ADAPT handles diverse locomotion tasks with two lightweight design choices. A key element is the adaptive sensing radius. It lets the policy regulate its sensing range online. In long-horizon tasks such as Stairs and Drops, the policy expands this range to anticipate terrain features. This behavior is interpretable and helps the agent plan stable foot placement ahead of times. 

Another key component is the vertical distance map $\mathbf{D}_t$. 
The compact 2D representation of 3D spatial constraints helps capture overhead and lateral obstacles that standard elevation maps miss in spatial-selection tasks such as Beams and Narrow Gates.
Compared with voxel-based methods such as Gallant, $\mathbf{D}_t$ is lighter and more noise-tolerant. It removes irrelevant details and keeps the policy focused on traversability. As a result, ADAPT maintains high success rates in real deployment, suggesting that well-structured inputs matter more than model complexity for agile humanoid navigation.

\section{CONCLUSION}
This paper presents ADAPT for a closed-loop integration of active perception and dynamic locomotion on humanoid robotics. The framework fuses a dual-projection environmental representation with an active sensing policy. This synthesis resolves the tension between high-fidelity spatial awareness and the strict computational limits of onboard systems. Crucially, we elevate the perceptual horizon from a static parameter to a dynamic control variable. This structural change enables the agent to autonomously achieve a trade-off between long-range anticipation and near-field precision. Physical deployment on the Unitree G1 humanoid robot validates the proposed approach. The robot achieved a $94.7\%$ success rate on diverse 3D terrains, with efficient computational resource allocation.

Future research will extend the active perception paradigm by incorporating simultaneous modulation of spatial resolution and sensor update frequency. We also aim to integrate semantic reasoning to support high-level navigation tasks. Enhancing predictive capabilities for dynamic agents will further advance safe and socially compliant humanoid operation in human-centric environments.

\balance
\bibliographystyle{IEEEtranBST/IEEEtran}
\bibliography{IEEEtranBST/IEEEabrv, main}

\clearpage
\appendix
\section{Training Details}

\subsection{PPO Hyperparameters}
The control policy was optimized via the Proximal Policy Optimization (PPO) algorithm. The hyperparameters governing the training process are enumerated in Table~\ref{tab:ppo_params}.

\begin{table}[h!]
\centering
\caption{PPO Hyperparameters}
\label{tab:ppo_params}
\begin{tabular}{lc}
\toprule
\textbf{Parameter} & \textbf{Value} \\
\midrule
Number of environments & 4096 \\
Number of steps per policy update & 24 \\
PPO epochs & 5 \\
Number of mini-batches & 4 \\
PPO clip parameter ($\epsilon$) & 0.2 \\
Value function clip parameter & 0.2 \\
GAE discount factor ($\lambda$) & 0.95 \\
Reward discount factor ($\gamma$) & 0.99 \\
Learning rate & 5e-4 \\
Entropy coefficient & 0.0 \\
Desired KL divergence & 0.01 \\
Max gradient norm & 1.0 \\
\bottomrule
\end{tabular}
\end{table}

\subsection{Policy Network Structure}
We utilize a recurrent actor-critic architecture leveraging a Gated Recurrent Unit (GRU) to handle temporal dependencies. The specific architectural configurations for the perceptual encoders and the recurrent policy layers are provided in Table~\ref{tab:network_structure}.

\begin{table}[h!]
\centering
\caption{Policy Network Structure}
\label{tab:network_structure}
\begin{tabular}{l|l|c}
\toprule
\multicolumn{2}{l|}{\textbf{Component}} & \textbf{Specification} \\
\midrule
\multirow{4}{*}{Recurrent Policy} & RNN Type & GRU \\
                           & RNN Hidden Dims & 256 \\
                           & Actor MLP Hidden Dims & [128, 64, 32] \\
                           & Critic MLP Hidden Dims & [128, 64, 32] \\
                           & Activation Function & ELU \\
\midrule
\multirow{3}{*}{Horizontal Map Encoder} & Encoder Type & MLP \\
                                 & MLP Hidden Dims & [256, 128, 64] \\
                                 & Activation Function & CELU \\
                                 & Output Embedding Dims & 32 \\
\midrule
\multirow{3}{*}{Vertical Map Encoder} & Encoder Type & MLP \\
                                 & MLP Hidden Dims & [128, 64] \\
                                 & Activation Function & CELU \\
                                 & Output Embedding Dims & 32 \\
\bottomrule
\end{tabular}
\end{table}

\subsection{Domain Randomization}
To facilitate robust simulation-to-reality transfer, we introduce stochastic variations across physical properties and sensor readings during training. The parameters and their respective randomization intervals are detailed in Table~\ref{tab:dr_params}.

\begin{table}[h!]
\centering
\caption{Parameters and Ranges for Domain Randomization}
\label{tab:dr_params}
\begin{tabular}{lc}
\toprule
\textbf{Parameter} & \textbf{Range (Uniform)} \\
\midrule
\multicolumn{2}{c}{Dynamics Randomization} \\
Robot Base Mass (kg) & [-1.0, 3.0] \\
Motor Strength Scale & [0.8, 1.2] \\
Ground Friction Coefficient & [0.1, 1.25] \\
\midrule
\multicolumn{2}{c}{External Perturbations} \\
Push Interval (s) & 5 \\
Max Push Velocity (m/s) & 1.5 \\
\midrule
\multicolumn{2}{c}{Sensor Randomization} \\
Proprioception Latency (s) & [0.005, 0.045] \\
Point Cloud Dropout & 10\% of points \\
LiDAR Point Noise & [-0.05, 0.05] \\
\bottomrule
\end{tabular}
\end{table}

\subsection{Reward Function Details}
The composite reward signal is a weighted summation of the components formulated in the primary text. The specific weights adopted for our finalized model are presented in Table~\ref{tab:reward_details}.

\begin{table}[h!]
\centering
\caption{Detailed Reward Component Weights}
\label{tab:reward_details}
\begin{tabular}{llc}
\toprule
\textbf{Category} & \textbf{Reward Term} & \textbf{Weight} \\
\midrule
\multirow{2}{*}{\textit{Task Fulfillment}} & Linear Velocity Tracking  & 2.0 \\
& Angular Velocity Tracking  & 0.5 \\
\midrule
\multirow{3}{*}{\textit{Locomotion Stability}} & Base Orientation  & -1.0 \\
& Base Angular Velocity XY  & -0.3 \\
& Hip Neutrality  & -1.0 \\
\midrule
\multirow{4}{*}{\textit{Behavior Regularization}} & Joint Velocity  & -1e-3 \\
& Joint Acceleration  & -2.5e-7 \\
& Action Rate  & -0.01 \\
& Energy  & -2.5e-7 \\
\midrule
\multirow{5}{*}{\textit{Safety and Survival}} & Horizontal Feet Collision  & -3.0 \\
& Mesh Penetration  & -1.0 \\
& Joint Limits  & -5.0 \\
& Alive Bonus  & 0.15 \\
& Foot Air Time  & 5.0 \\
\midrule
\multirow{3}{*}{\textit{Adaptive Perception}} & Radius Adaptive & 1.0 \\
& Radius Smoothness & -0.1 \\
& Radius Regularization & -0.1 \\
\bottomrule
\end{tabular}
\end{table}



\end{document}